\documentclass[10pt,twocolumn,letterpaper]{article} 

\usepackage{avss}
\usepackage{times}
\usepackage{epsfig}
\usepackage{graphicx}
\usepackage{amsmath}
\usepackage{amssymb}

\usepackage{kotex}

\newcommand{\yj}[1]{}
\newcommand{\dw}[1]{}

\avssfinalcopy 

\def\avssPaperID{182} 

\ifavssfinal\fi
\begin{document}

\title{Distributional Uncertainty for Out-of-Distribution Detection}

\author{JinYoung Kim$^{1}$, DaeUng Jo$^{2}$, Kimin Yun$^{3,4}$, Jeonghyo Song$^{1}$, Youngjoon Yoo$^{1,*}$\\
$^1$Department of Artificial Intelligence, Chung-Ang University, Seoul, Korea\\
$^2$School of Electronics Engineering, Kyungpook National University, Daegu, Korea\\
$^3$Visual Intelligence Lab., ETRI, Daejeon, Korea\\
$^4$University of Science and Technology~(UST), Daejeon, Korea\\
{\tt\small \{barraki7226, thd9592s\}@cau.ac.kr, daeung.jo@knu.ac.kr, kimin.yun@etri.re.kr, yjyoo3312@gmail.com}
\thanks{Corresponding author}
}
\maketitle

\begin{abstract} 
Estimating uncertainty from deep neural networks is a widely used approach for detecting out-of-distribution (OoD) samples, which typically exhibit high predictive uncertainty. However, conventional methods such as Monte Carlo (MC) Dropout often focus solely on either model or data uncertainty, failing to align with the semantic objective of OoD detection. To address this, we propose the Free-Energy Posterior Network, a novel framework that jointly models distributional uncertainty and identifying OoD and misclassified regions using free energy. Our method introduces two key contributions: (1) a free-energy-based density estimator parameterized by a Beta distribution, which enables fine-grained uncertainty estimation near ambiguous or unseen regions; and (2) a loss integrated within a posterior network, allowing direct uncertainty estimation from learned parameters without requiring stochastic sampling. By integrating our approach with the residual prediction branch (RPL) framework, the proposed method goes beyond post-hoc energy thresholding and enables the network to learn OoD regions by leveraging the variance of the Beta distribution, resulting in a semantically meaningful and computationally efficient solution for uncertainty-aware segmentation.
We validate the effectiveness of our method on challenging real-world benchmarks, including Fishyscapes, RoadAnomaly, and Segment-Me-If-You-Can.
\end{abstract}

\section{Introduction}

In safety-critical applications such as autonomous driving, semantic segmentation models must not only produce accurate predictions but also estimate uncertainty to identify unreliable or out-of-distribution (OoD) regions~\cite{kendall2017uncertainties}. A central challenge in this context is to detect anomalous inputs that deviate from the training distribution and often lead to overconfident errors. One widely used approach, Monte Carlo (MC) Dropout~\cite{gal2016dropout}, estimates model uncertainty by performing multiple stochastic forward passes. However, it suffers from high computational overhead and may not reliably reflect semantic uncertainty~\cite{mukhoti2018evaluating}.
\begin{figure}[t]
    \centering
\includegraphics[width=0.95\linewidth]{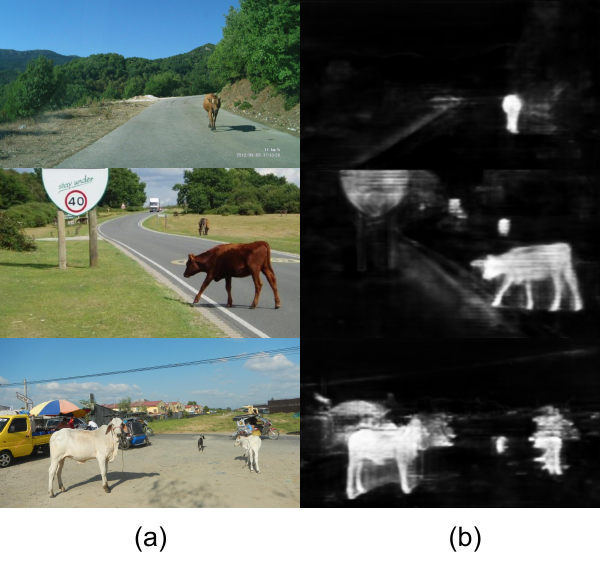}
\caption{
Comparison on the Road Anomaly dataset. (a) Input images. (b) Results from our proposed method (RPL + Ours).
}

    \label{fig:teaser}
\end{figure}

To address these limitations, recent research has focused on estimating \textit{distributional uncertainty}, which aims to capture ambiguity stemming from unseen or ambiguous data rather than from model stochasticity~\cite{malinin2018predictive}. Posterior Networks~\cite{charpentier2020posterior} and evidential deep learning~\cite{sensoy2018evidential} have introduced the use of Dirichlet and Beta distributions to model predictive uncertainty, providing richer and more interpretable confidence estimates. In semantic segmentation, Siddharth et al.~\cite{ancha2024deep} extended this idea to the pixel level, enabling fine-grained uncertainty estimation for OoD detection. However, these approaches often require additional sampling, post-hoc thresholding, or hand-crafted labels for OoD regions, which may hinder scalability and generalization.
Another line of work leverages free-energy-based confidence measures for OoD detection. FlowEneDet~\cite{gudovskiy2023concurrent} introduced a flow-based density estimator that identifies inlier and outlier regions without explicit supervision, using free energy as a surrogate for likelihood. Building on this, Residual Pattern Learning (RPL)~\cite{liu2023residual} achieved state-of-the-art OoD detection performance by penalizing regions with high free energy using contrastive learning. However, RPL relies on fixed free-energy targets and post-hoc thresholds, limiting its flexibility, especially in early training stages.

\begin{figure*}[t]
    \centering
    \includegraphics[width=0.95\textwidth]{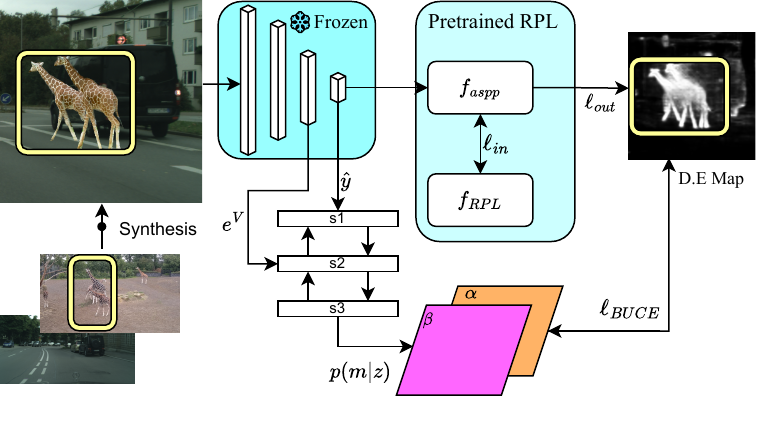}
    \caption{
    Overview of the proposed architecture. Outlier Exposure (OE) samples are generated by synthesizing inlier (Cityscapes) and outlier (COCO) images, then passed through a frozen segmentation backbone into two branches: the RPL framework and a flow-based model (FlowEneDet). The flow model uses coupling blocks ($s_1$, $s_2$, $s_3$) to compute free energy representations, where $e^v$ denotes the ReLU-activated feature and $\hat{y}$ is extracted via the conv1 block. These representations are transformed into Beta posterior parameters $(\alpha, \beta)$. A differential entropy (D.E) map is then derived and replaces the free energy term in RPL’s $\ell_{\text{in}}$ loss, enabling more accurate OoD detection.
    }

    
    \label{fig:architecture}
\end{figure*}

To overcome these challenges, we propose a unified framework that combines the strengths of both paradigms. Specifically, we integrate a frozen segmentation model with an additional learnable flow-based posterior network that estimates a Beta distribution at each pixel. From these Beta parameters, we compute the variance as a direct measure of distributional uncertainty, which is injected into the training process via a novel Beta Uncertainty Cross Entropy and Energy (BUCE) loss. This variance-based signal allows the model to emphasize ambiguous OoD regions while suppressing overconfident predictions in known areas, leading to more stable and semantically grounded uncertainty-aware learning.
Figure~\ref{fig:architecture} illustrates the overall framework. Compared to prior work, our approach enables end-to-end uncertainty estimation and OoD segmentation without relying on sampling or handcrafted thresholds, resulting in a more principled and efficient solution.
We validate the effectiveness of the proposed framework through extensive experiments on real-world OoD benchmarks, including Fishyscapes~\cite{fishyscapes} (Static and LostAndFound), RoadAnomaly~\cite{roadanomaly}, and Segment-Me-If-You-Can (SMIYC)~\cite{smiyc}.

\section{Preliminaries}

\subsection{Uncertainty Posterior Network}

In evidential deep learning, predictive uncertainty is decomposed into three components: \emph{data}, \emph{distributional}, and \emph{model} uncertainty. These reflect variability in the input, the latent representation, and the learned parameters, respectively. The overall uncertainty can be estimated in a single forward pass via the posterior framework in Equation~\ref{eq:prior}:

{\footnotesize
\begin{equation}
\mathbb{P}(\omega_c \mid x^*, \mathcal{D}) = 
\int \!\!\! \int 
\underbrace{\mathbb{P}(\omega_c \mid \mu)}_{\text{Data}} \,
\underbrace{\mathbb{P}(\mu \mid x^*, \theta)}_{\text{Distributional}} \,
\underbrace{\mathbb{P}(\theta \mid \mathcal{D})}_{\text{Model}} \,
d\mu\, d\theta.
\label{eq:prior}
\end{equation}
}

Here, $\omega_c$ is the class label, $x^*$ the test input, and $\mathcal{D}$ the training dataset. Posterior networks approximate this distribution using a Dirichlet formulation, allowing class-wise predictions with associated confidence levels.

To improve tractability, normalizing flows are used to transform inputs into a latent space where the Dirichlet parameters $\boldsymbol{\alpha}$ are defined as follows:

{\footnotesize
\begin{equation}
\mathbb{E}_{\mathbf{p} \sim \text{Dir}(\boldsymbol{\alpha}^{(i)})} [p_c] = 
\frac{\beta_c^{\text{prior}} + N \cdot P(c \mid \mathbf{z}^{(i)}; \boldsymbol{\phi}) \cdot P(\mathbf{z}^{(i)}; \boldsymbol{\phi})}
{\sum_{c} \beta_c^{\text{prior}} + N \cdot P(\mathbf{z}^{(i)}; \boldsymbol{\phi})}.
\label{eq:dir_expected}
\end{equation}
}

Here, $P(c \mid \mathbf{z}^{(i)}; \boldsymbol{\phi})$ denotes the predicted class probability for latent feature $\mathbf{z}^{(i)}$, and $N$ is a confidence scaling factor.

In binary classification, the Dirichlet reduces to the Beta distribution, with $\alpha$ and $\beta$ representing soft evidence for inlier and outlier classes. The expected inlier probability is:

{\footnotesize
\begin{equation}
\mathbb{E}[p_{\text{in}}] = 
\frac{P_{\text{in}}(\mathbf{z}^{(i)}; \boldsymbol{\phi})}
{P_{\text{in}}(\mathbf{z}^{(i)}; \boldsymbol{\phi}) + P_{\text{out}}(\mathbf{z}^{(i)}; \boldsymbol{\phi})}
\label{eq:beta_expected}
\end{equation}
}

The corresponding uncertainty is quantified via the Beta variance:

\begin{equation}
\text{Var}(p) = \frac{\alpha \beta}{(\alpha + \beta)^2 (\alpha + \beta + 1)}.
\label{eq:beta_variance}
\end{equation}

\subsection{Energy-based Flow Model}

We adopt a flow-based density estimation framework in which the negative log-likelihood is interpreted as a free energy term. Normalizing flows provide a flexible mechanism to map complex data distributions into a tractable latent space through a sequence of invertible transformations, enabling exact and expressive density estimation.

Given a latent variable $z$ obtained through such a transformation, we model its density under a Gaussian distribution. The corresponding log-likelihood is:

\begin{equation}
\log \mathcal{N}(u \mid \mu, \Sigma) = \sum_{d=1}^{D} \text{diag}(U)_d - \frac{1}{2} \left\| U(z - \mu) \right\|_2^2,
\label{eq:flow_energy}
\end{equation}

where $U$ is the Cholesky decomposition of the inverse covariance matrix $\Sigma^{-1}$, and $\mu$ is the Gaussian mean. The quadratic term penalizes deviations from the mean, while the diagonal term accounts for the learned scale in each dimension. In this view, low-energy regions correspond to familiar, high-confidence inputs, while high-energy values indicate out-of-distribution (OoD) samples.

This interpretation is particularly useful for semantic segmentation tasks, where dense, spatially resolved uncertainty maps are required. Evaluating energy across the feature map allows identification of uncertain or ambiguous regions.

We also compute class-conditional likelihoods via Bayes’ rule to support discriminative prediction:

\begin{equation}
p_\theta(m \mid z) = \frac{p_\theta(z \mid m)p(m)}{\sum_{d=1}^{D} p_\theta(z \mid m = d)},
\label{eq:flow_bayes}
\end{equation}

where $p_\theta(z \mid m)$ is the likelihood of $z$ under class $m$, and $p(m)$ is the prior. This enables per-class density modeling and uncertainty estimation based on relative likelihoods.

Overall, the energy-based flow model complements posterior networks by providing an explicit likelihood-based view of uncertainty and anomaly.

\section{Proposed Method}

In this section, we present our proposed method for robust out-of-distribution (OoD) detection in semantic segmentation. The overall framework is composed of two major components:  (1) a flow-based posterior network to estimate predictive uncertainty using the Beta distribution, (2) a free-energy-based segmentation model with a residual prediction branch (RPL) with a unified training objective that integrates uncertainty-aware loss functions.

\subsection{Uncertainty Formulation}
Figure~\ref{fig:architecture} illustrates the overall architecture. Given an input image, the model first computes free energy scores via a standard segmentation backbone. These features are then passed to a flow-based network that estimates the parameters $\alpha$ and $\beta$ of the Beta distribution.
Specifically, given the latent representation $\mathbf{z}^{(i)}$ of pixel $i$, the flow-based network outputs a log-probability vector $\mathbf{z}^{(i)} = [z_{\text{in}}, z_{\text{out}}]$, converted into Beta parameters as:
\begin{eqnarray}
\begin{aligned}
\alpha^{(i)} &= 1 + \log(1 + e^{z_{\text{in}}^{(i)}}),\\
\beta^{(i)} &= 1 + \log(1 + e^{z_{\text{out}}^{(i)}}).
\label{eq:beta_param}   
\end{aligned}    
\end{eqnarray}
From these parameters, we compute the expected inlier confidence score $\bar{p}^{(i)}$ by the mean of the Beta distribution:
\begin{equation}
\bar{p}^{(i)} = \mathbb{E}[p^{(i)}] = \frac{\alpha^{(i)}}{\alpha^{(i)} + \beta^{(i)}}.
\label{eq:beta_expectation}
\end{equation}
The hard label prediction $y^{(i)}$ for pixel $i$ is conducted by thresholding or using $\arg\max$ value between inlier and outlier scores, respectively:

\begin{equation}
y^{(i)} = \begin{cases}
1, & \text{if } \bar{p}^{(i)} \geq \tau \\
0, & \text{otherwise}.
\end{cases}
\label{eq:beta_prediction}
\end{equation}

To capture distributional uncertainty, our method leverages a flow-based posterior network to estimate the parameters of a Beta distribution at each pixel. Specifically, given an input feature \( z \) derived from free-energy-based representations, the posterior network outputs pixel-wise \( \alpha \) and \( \beta \):
\begin{eqnarray}
\begin{aligned}
\alpha(\omega) &= 1 + \log P(z(\omega)|c), \\
\beta(\omega) &= 1 + \log Q(z(\omega)|c).
\end{aligned}
\end{eqnarray}
Here, \( P \) and \( Q \) denote the flow-based probabilistic mappings conditioned on contextual features \( c \), and \( \omega \) denotes the spatial location.
We then compute the predictive variance using the Beta distribution:
\begin{equation}
\text{Var}(p(\omega)) = \frac{\alpha(\omega)\beta(\omega)}{(\alpha(\omega)+\beta(\omega))^2(\alpha(\omega)+\beta(\omega)+1)}.
\label{eq:beta_variance}
\end{equation}
Finally, to encourage the model to emphasize uncertain OoD regions, this variance is directly injected into the energy-based loss term of the RPL framework:
\begin{equation}
\mathcal{L}_{\text{out}} = \sum_{\omega \in \Omega} \max\left(-\mathbf{m}(\omega) \cdot \text{Var}(p(\omega)),\ 0\right),
\label{eq:loss_out_var}
\end{equation}
where \( \mathbf{m}(\omega) \in \{0, 1\} \) is a binary mask indicating OoD pixels.
We note that, since \( \alpha \) and \( \beta \) are generated by the flow model, minimizing \( \mathcal{L}_{\text{out}} \) induces gradient updates that propagate back to the flow parameters via the variance term:
\begin{equation}
\nabla_{\theta_{\text{flow}}} \mathcal{L}_{\text{out}} = 
- \frac{\partial \text{Var}(p)}{\partial \alpha} \cdot \frac{\partial \alpha}{\partial \theta_{\text{flow}}}
- \frac{\partial \text{Var}(p)}{\partial \beta} \cdot \frac{\partial \beta}{\partial \theta_{\text{flow}}}.
\end{equation}
This encourages the flow network to produce Beta parameters with higher variance for OoD regions. Since the variance of a Beta distribution is maximized when \( \alpha \approx \beta \) and both are small, the model implicitly learns to represent uncertainty by reducing the sharpness of the predictive distribution in OoD areas. This dynamic adaptation facilitates robust discrimination between inlier and outlier regions and enhances the stability of RPL training from early stages.

\subsection{Free-energy-based Segmentation Model}
To incorporate uncertainty into the supervision signal, we design a unified loss function that accounts for both inlier confidence and outlier ambiguity using the Beta posterior. The proposed Beta-Uncertainty aware Cross Entropy(BUCE) loss consists of three components: 
\paragraph{(1) Standard RPL Cross Entropy.}  
We first compute the classical cross-entropy loss between the predicted logit \( \hat{y} \) and the pseudo-target \( \tilde{y} \) generated from vanilla logits:
\begin{equation}
\mathcal{L}_{\text{ce}} = \text{CE}(\hat{y},\ \tilde{y}) = - \sum_{c} \tilde{y}_c \log \hat{y}_c.
\end{equation}
\paragraph{(2) Beta Uncertainty Cross Entropy (UCE).}
To regulate the learning signal in ambiguous or unknown regions, we introduce the Beta-based Uncertainty Cross Entropy Loss (BUCE) based on UCE loss\cite{landgraf2023u}. This term penalizes the mismatch between model confidence and the uncertainty modeled by the Beta posterior.
\begin{equation}
\begin{split}
\mathcal{L}_{\text{UCE}} = \sum_{i} y_i \cdot \left[ 
    \psi(\alpha_i + \beta_i) - \psi(\alpha_i) - \psi(\beta_i) \right] \\
    - \lambda_{\text{reg}} \cdot \mathbb{H}\left[\text{Beta}(\alpha_i, \beta_i)\right].
\end{split}
\label{eq:uce_loss}
\end{equation}
Here, $\psi(\cdot)$ is the digamma function and $\mathbb{H}[\text{Beta}(\alpha, \beta)]$ denotes the differential entropy of the Beta distribution:
\begin{equation}
\begin{split}
\mathbb{H}[\text{Beta}(\alpha, \beta)]& =\ 
 \ln B(\alpha, \beta) 
- (\alpha - 1) \psi(\alpha) \\
& - (\beta - 1) \psi(\beta) 
+ (\alpha + \beta - 2) \psi(\alpha + \beta).
\end{split}
\label{eq:beta_entropy}
\end{equation}

The first term in Equation~\ref{eq:uce_loss} provides a Cross Entropy signal based on the prediction confidence. The entropy term encourages higher uncertainty in ambiguous regions, helping the model avoid overconfident errors. This regularization is especially important for representing distributional uncertainty near OoD boundaries. In practice, we scale $\mathcal{L}_{\text{UCE}}$ by $10^{-7}$ for numerical stability.
\paragraph{(3) Beta Variance Consistency.}  
We enforce consistency between the predicted variance of the Beta distribution and the binary OoD target map using binary cross-entropy:
\begin{equation}
\text{Var}^{(i)} = \frac{\alpha^{(i)} \beta^{(i)}}{(\alpha^{(i)} + \beta^{(i)})^2(\alpha^{(i)} + \beta^{(i)} + 1)}.
\end{equation}
\begin{equation}
\mathcal{L}_{\text{var}} = \text{BCE}(\text{Var}^{(i)},\ y^{(i)}),
\end{equation}

\paragraph{(Final)}  
The final BUCE loss is defined as:
\begin{equation}
\mathcal{L}_{\text{BUCE}} = \mathcal{L}_{\text{ce}} + \lambda_1 \cdot \mathcal{L}_{\text{uce}} + \lambda_2 \cdot \mathcal{L}_{\text{var}},
\end{equation}
where \( \lambda_1, \lambda_2 \) are hyperparameters controlling the influence of uncertainty-aware terms.



\section{Experiments}

\begin{table*}[t]
\centering
\caption{Comparison on Fishyscapes and SMIYC validation sets. Best results are in \textbf{bold}.}
\resizebox{0.99\textwidth}{!}{
\begin{tabular}{l|ccc|ccc|ccc|ccc}
\hline
{Method} & 
\multicolumn{3}{c|}{Fishyscapes - Static} & 
\multicolumn{3}{c|}{Fishyscapes - L\&F} & 
\multicolumn{3}{c|}{SMIYC - Anomaly} & 
\multicolumn{3}{c}{SMIYC - Obstacle} \\
& FPR↓ & AuPRC↑ & AUROC↑ & FPR↓ & AuPRC↑ & AUROC↑ & FPR↓ & AuPRC↑ & AUROC↑ & FPR↓ & AuPRC↑ & AUROC↑ \\
\hline
MCD(RN101)~\cite{liu2023residual}       & {25.20} & 19.43 & {92.12} & {30.74} & 6.62 & 90.97 & 57.75 & 45.94 & 81.89 & {11.52} & {48.22} & 97.46 \\
SE(RN101)~\cite{zhao2024segment}  &25.22 & 19.68 & 92.12 & 29.59 & 7.04 & 91.19 & 58.18& 45.33 & 81.77 & 11.92 & 48.41 &97.43  \\
EnE(RN101)                             & 15.81 &  36.61& 96.24 & \textbf{26.61} & \textbf{12.56} & \textbf{94.31} & 54.32 & 43.25 & 82.94 & 15.20 & 52.01& 97.31 \\
\textbf{Ours + DE(RN101)}~\cite{liu2023residual}        & \textbf{13.23} & \textbf{56.34} & \textbf{97.26} & {26.68} & 11.56 & {94.19}   & \textbf{48.09} & \textbf{56.55} & \textbf{86.98} & \textbf{11.08} &\textbf{52.42}  & \textbf{98.01} \\

\hline
\end{tabular}
}
\label{tab:fs_smiyc_table}
\end{table*}

\begin{table}[t]
\centering
\caption{\textbf{RoadAnomaly validation set.} 
All approaches are based on the \textbf{DeepLabv3+} architecture and best results are in \textbf{boldface}.}
\label{tab:roadanomaly_sota}

\resizebox{\linewidth}{!}{
\begin{tabular}{|l|c|c|c|}
\hline
\textbf{Method} & FPR~$\downarrow$ & AuPRC~$\uparrow$ & AuROC~$\uparrow$ \\
\hline
SE (ResNet101) & 73.93 & 19.24 & 69.49 \\
MCD (ResNet101) & 73.25 & 19.20 & 69.66 \\
ENE (ResNet101) & 74.13 & 19.15 & 70.56 \\
\textbf{Ours + DE (ResNet101)} & \textbf{71.96} & \textbf{19.76} & 70.73 \\
\textbf{Ours + DE (MobileNet)} & 72.74 & 18.82 & \textbf{71.19} \\
\hline
GMMSeg & 47.90 & 34.42 & 84.71 \\
PEBAL & 44.58 & 45.10 & 87.63 \\
RPL+CoroCL & \textbf{17.74} & 71.60 & 95.72 \\
\textbf{RPL+CoroCL+DE (Ours)} & 18.49 & \textbf{75.49} & \textbf{95.83} \\
\hline
\end{tabular}
}
\end{table}

 
\begin{table}[t]
\centering
\caption{Performance comparison of RPL variants on Fishyscapes and SMIYC validation sets in terms of FPR and AUPRC.}
\vspace{2mm}
\label{tab:rpl_ap_auprc}
\resizebox{\linewidth}{!}{%
\begin{tabular}{lcccc}
\hline
\textbf{Method} & \multicolumn{2}{c}{{Fishyscapes - Static}} & \multicolumn{2}{c}{{Fishyscapes - L\&F}} \\
                & \textbf{FPR}$\downarrow$     & \textbf{AUPRC}$\uparrow$ & \textbf{FPR}$\downarrow$   &\textbf{AUPRC}$\uparrow$ \\
\hline

RPL + SE        & 6.59  & 89.53 & 16.88 & 67.36 \\
PEBAL          & 1.52  & 92.08 & 4.76 & 58.81 \\
DenseHybrid     & 4.17  & 76.23 & 5.09 &69.79  \\
RPL            & \textbf{0.85}  & 92.46 & 4.76  & {70.61} \\

\textbf{RPL + DE (Ours)} & 1.03  & \textbf{93.77} & \textbf{3.53}  & \textbf{71.28} \\
\hline
 & \multicolumn{2}{c}{{SMIYC - Anomaly}} & \multicolumn{2}{c}{{SMIYC - Obstacle}} \\
                & \textbf{FPR}$\downarrow$   & \textbf{AUPRC}$\uparrow$ & \textbf{FPR}$\downarrow$    & \textbf{AUPRC}$\uparrow$ \\
\hline
RPL + SE        & 22.36  & 86.21 & 0.57  & 92.59 \\
PEBAL           & 36.74  & 53.10 & 7.92 & 10.45 \\
DenseHybrid     & 52.65  & 61.08 & 0.71 &89.49  \\
RPL                & 7.18  & 88.55 & \textbf{0.09}  & \textbf{96.91} \\
\textbf{RPL + DE (Ours)} & \textbf{6.87}  & \textbf{89.48} & {0.11}  & {96.29} \\
\hline
\end{tabular}%
}
\end{table}

In this section, we first explain the experimental setup, and compare our method across various baselines while analyzing the relationship between the Beta distribution parameters $\alpha$ and $\beta$ in the OoD context.

\subsection{Experimental Setting}
We follow the standard evaluation protocol of the RPL framework~\cite{liu2023residual}. The model is trained on the Cityscapes~\cite{cityscapes} dataset (2,975 training / 500 validation images), with 46,751 outlier exposure (OE) samples from COCO~\cite{coco}, ensuring no label overlap.
We evaluate the proposed method on four OoD segmentation benchmarks: Fishyscapes~\cite{fishyscapes} (Static, LostAndFound), RoadAnomaly~\cite{roadanomaly}, and Segment-me-If-You-Can (SMIYC)~\cite{smiyc} dataset for both Anomaly and Obstacle tracks, including SMIYC-L\&F, a cleaned subset of LostAndFound.
All experiments use a frozen DeepLabV3+ backbone~\cite{deeplab}, with input resized to $700 \times 700$ for efficiency. Metrics include FPR95, AUROC, AP, and AUPRC, assessing separability, robustness, and OoD localization.
\begin{figure*}[t]
    \centering
    \includegraphics[width=\linewidth]{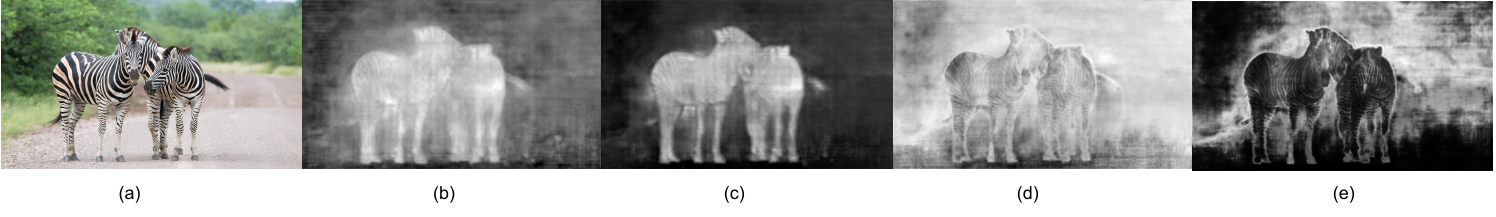} \\
\caption{
Qualitative comparison of uncertainty maps on the Road Anomaly dataset. (a) Input image, (b) baseline RPL~\cite{liu2023residual}, (c) our proposed method using differential entropy from the Beta posterior, (d) RPL with Shannon entropy, and (e) RPL with MC Dropout~\cite{gal2016dropout} variance map. The visualization demonstrates that our method more effectively isolates out-of-distribution (OoD) regions while significantly reducing responses on inlier areas. Compared to other uncertainty estimation methods, our approach provides sharper, more localized anomaly boundaries and better focuses on true OoD pixels, minimizing distractions from seen or ambiguous regions.
}

\label{fig:qual}
\end{figure*}
\subsection{Experimental Results}
We compare our proposed method with several uncertainty estimation baselines on four major OoD segmentation benchmarks, and summarize the results in Table~\ref{tab:fs_smiyc_table}, Table~\ref{tab:roadanomaly_sota}, and Table~\ref{tab:rpl_ap_auprc}. All methods are based on the same backbone, DeepLabV3+ with ResNet101 (RN101), to ensure fairness. In Table~\ref{tab:fs_smiyc_table}, we evaluate against representative methods such as Shannon Entropy (SE), Monte Carlo Dropout (MCD), and Free Energy (EnE). 
Our method (Ours + Differential Entropy) outperforms all baselines in terms of FPR and AuPRC across most datasets. For instance, on Fishyscapes-Static, our approach achieves the lowest FPR (13.23) and the highest AuPRC (56.34), significantly outperforming MCD (FPR 25.20, AuPRC 19.43) and EnE (FPR 15.81, AuPRC 36.61). Similarly, on the SMIYC-Anomaly benchmark, our method achieves the best FPR (48.09) and AuPRC (56.55), indicating improved robustness under open-set conditions. 
Table~\ref{tab:roadanomaly_sota} shows results on the RoadAnomaly dataset. Compared to strong baselines like RPL+CoroCL, our method achieves the best AuPRC (75.49) and maintains a low FPR (18.49), validating that our Beta-based variance map provides meaningful supervision. 

Finally, in Table~\ref{tab:rpl_ap_auprc}, we conduct an ablation comparing RPL variants. Compared to RPL alone or RPL+SE, our RPL+DE consistently improves both AP and AUPRC, especially in difficult datasets like SMIYC-Anomaly (AP 89.48 vs. 57.99) and Obstacle (AUPRC 96.29 vs. 22.10), confirming the effectiveness of our distributional uncertainty.

\subsection{Qualitative Analysis}
As shown in Figure~\ref{fig:qual}, we compare the inference results between various uncertainty estimation methods and our proposed approach on the Fishyscapes dataset. The visualization demonstrates that our method more effectively highlights anomalous objects in the scene while suppressing responses in known areas. By incorporating distributional uncertainty, our approach produces sharper and more precise anomaly boundaries, resulting in clearer out-of-distribution (OoD) pixel detection compared to the baselines. This enhanced detection capability is particularly evident in how our model emphasizes OoD object regions against the background, which is critical for autonomous driving systems to identify potential road hazards.

\section{Conclusion}
In this paper, we proposed a novel approach to capture pure distributional uncertainty by leveraging free energy through a flow-based model, FlowEneDet, in conjunction with a Beta posterior network. This framework allows for effective estimation of uncertainty in unknown data regions. Furthermore, we integrated our method into the state-of-the-art RPL framework, using the variance map derived from the Beta posterior as a principled supervisory signal. This integration enables the model to more robustly guide learning in OoD regions, grounded in distributional uncertainty rather than fixed energy thresholds. Future work includes integrating our uncertainty map into diverse energy-based models to assess its general applicability in OoD detection.

\section{Acknowledgment}
This work was supported by the Institute of Information $\&$ Communications Technology Planning $\&$ Evaluation (IITP) grant funded by the Korea government (MSIT) [RS-2021-II211341, Artificial Intelligence Graduate School Program (Chung-Ang University) and RS-2022-II220124, Development of Artificial Intelligence Technology for Self-Improving Competency-Aware Learning Capabilities].

{\small
\bibliographystyle{ieee}
\bibliography{egbib}

\begin{thebibliography}{10}\itemsep=-1pt

\bibitem{ancha2024deep}
S.~Ancha, P.~R. Osteen, and N.~Roy.
\newblock Deep evidential uncertainty estimation for semantic segmentation under out-of-distribution obstacles.
\newblock In {\em 2024 IEEE International Conference on Robotics and Automation (ICRA)}, pages 6943--6951. IEEE, 2024.

\bibitem{fishyscapes}
H.~Blum, P.-E. Sarlin, J.~Nieto, R.~Siegwart, and C.~Cadena.
\newblock The fishyscapes benchmark: Measuring blind spots in semantic segmentation.
\newblock {\em arXiv preprint arXiv:1904.03215}, 2019.

\bibitem{smiyc}
R.~Chan, K.~Lis, S.~Uhlemeyer, H.~Blum, S.~Honari, R.~Siegwart, P.~Fua, M.~Salzmann, and M.~Rottmann.
\newblock Segmentmeifyoucan: A benchmark for anomaly segmentation.
\newblock In {\em Thirty-fifth Conference on Neural Information Processing Systems Datasets and Benchmarks Track}, 2021.

\bibitem{charpentier2020posterior}
B.~Charpentier, D.~Z{\"u}gner, and S.~G{\"u}nnemann.
\newblock Posterior network: Uncertainty estimation without ood samples via density-based pseudo-counts.
\newblock {\em Advances in neural information processing systems}, 33:1356--1367, 2020.

\bibitem{deeplab}
L.-C. Chen, Y.~Zhu, G.~Papandreou, F.~Schroff, and H.~Adam.
\newblock Encoder-decoder with atrous separable convolution for semantic image segmentation.
\newblock In {\em Proceedings of the European conference on computer vision (ECCV)}, pages 801--818, 2018.

\bibitem{cityscapes}
M.~Cordts, M.~Omran, S.~Ramos, T.~Rehfeld, M.~Enzweiler, R.~Benenson, U.~Franke, S.~Roth, and B.~Schiele.
\newblock The cityscapes dataset for semantic urban scene understanding.
\newblock In {\em Proceedings of the IEEE conference on computer vision and pattern recognition}, pages 3213--3223, 2016.

\bibitem{gal2016dropout}
Y.~Gal and Z.~Ghahramani.
\newblock Dropout as a bayesian approximation: Representing model uncertainty in deep learning.
\newblock {\em ICML}, 2016.

\bibitem{gudovskiy2023concurrent}
D.~Gudovskiy, T.~Okuno, and Y.~Nakata.
\newblock Concurrent misclassification and out-of-distribution detection for semantic segmentation via energy-based normalizing flow.
\newblock In {\em Uncertainty in Artificial Intelligence}, pages 745--755. PMLR, 2023.

\bibitem{kendall2017uncertainties}
A.~Kendall and Y.~Gal.
\newblock What uncertainties do we need in bayesian deep learning for computer vision?
\newblock In {\em NeurIPS}, 2017.

\bibitem{landgraf2023u}
S.~Landgraf, M.~Hillemann, K.~Wursthorn, and M.~Ulrich.
\newblock U-ce: Uncertainty-aware cross-entropy for semantic segmentation.
\newblock {\em arXiv preprint arXiv:2307.09947}, 2023.

\bibitem{coco}
T.~Y. Lin, M.~Maire, S.~Belongie, J.~Hays, P.~Perona, D.~Ramanan, P.~Dollar, and C.~L. Zitnick.
\newblock {Microsoft COCO: Common objects in context}.
\newblock In {\em ECCV}, 2014.

\bibitem{roadanomaly}
K.~Lis, K.~Nakka, P.~Fua, and M.~Salzmann.
\newblock Detecting the unexpected via image resynthesis.
\newblock In {\em Proceedings of the IEEE/CVF International Conference on Computer Vision}, pages 2152--2161, 2019.

\bibitem{liu2023residual}
Y.~Liu, C.~Ding, Y.~Tian, G.~Pang, V.~Belagiannis, I.~Reid, and G.~Carneiro.
\newblock Residual pattern learning for pixel-wise out-of-distribution detection in semantic segmentation.
\newblock In {\em Proceedings of the IEEE/CVF International Conference on Computer Vision}, pages 1151--1161, 2023.

\bibitem{malinin2018predictive}
A.~Malinin and M.~Gales.
\newblock Predictive uncertainty estimation via prior networks.
\newblock {\em Advances in neural information processing systems}, 31, 2018.

\bibitem{mukhoti2018evaluating}
J.~Mukhoti and Y.~Gal.
\newblock Evaluating bayesian deep learning methods for semantic segmentation.
\newblock {\em arXiv preprint arXiv:1811.12709}, 2018.

\bibitem{sensoy2018evidential}
M.~Sensoy, L.~Kaplan, and M.~Kandemir.
\newblock Evidential deep learning to quantify classification uncertainty.
\newblock {\em Advances in neural information processing systems}, 31, 2018.

\bibitem{zhao2024segment}
W.~Zhao, J.~Li, X.~Dong, Y.~Xiang, and Y.~Guo.
\newblock Segment every out-of-distribution object.
\newblock In {\em Proceedings of the IEEE/CVF Conference on Computer Vision and Pattern Recognition}, pages 3910--3920, 2024.

\end{thebibliography}
}

\end{document}